# How to choose features to improve prediction performance in lane-changing intention: A meta-analysis

**Ruifeng Gu, Research Assistant**

School of Traffic and Transportation Engineering, Central South University, Changsha, Hunan 410075, P. R. China
Email: guruifeng@knights.ucf.edu

**Abstract:**
Lane-change is a fundamental driving behavior and highly associated with various types of collisions, such as rear-end collisions, sideswipe collisions, and angle collisions and the increased risk of a traffic crash. This study investigates effectiveness of different features categories combination in lane-changing intention prediction. Studies related to lane-changing intention prediction have been selected followed by strict standards. Then the meta-analysis was employed to not only evaluate the effectiveness of different features categories combination in lane-changing intention but also capture heterogeneity, effect size combination, and publication bias. According to the meta-analysis and reviewed research papers, results indicate that using input features from different types can lead to different performances. And vehicle input type has a better performance in lane-changing intention, prediction, compared with environment or even driver combination input type. Finally, some potential future research directions are proposed based on the findings of the paper.

## 1. INTRODUCTION

According to the traffic safety report published by National Highway Traffic Safety Administration (NHTAS,2019), there were 36,560 people killed in motor vehicle traffic crashes on U.S. roadways during 2018. And the critical reason, which is the last event in the crash causal chain, was assigned to the driver in 94 percent of the investigated crashes (Singh, 2015).

Many efforts have been devoted to preventing the accident from happening by reducing drivers' workloads and decision errors, such as the side warning assistance (SWA) of ADAS even Higher-level autonomous driving technology techniques. The revision of SAE (The Society of Automotive Engineers) J3016 has further clarified the differences between SAE Level 3 and SAE Level 4, including the role of the fallback-ready user, the possibility of some automated fallback at SAE Level 3, and the possibility of some alerts to in-vehicle users at SAE Level 4. The driver must take over and drive when the features request in Level 3 autonomous. And Level 4 autonomous driving features will not require you to take over driving. Apollo developed by Baidu, which is one of the most intelligent autonomous vehicles in the world, just reached Level 3 autonomous.

Lane-change, a fundamental and complex driving behavior, is highly associated with various types of collisions, such as rear-end collisions, sideswipe collisions, and angle collisions (Ahn et al., 2010) and the increased risk of a traffic crash (Pande & Abdel-Aty, 2006). For example, the number of motor vehicle crashes due to sideswipe (related to lane-change) is approximately 863,100 in the United States in 2018 (NHTSA, 2019), among which a large proportion of crashes are caused by human error. Therefore, understanding driver intention is beneficial to traffic safety and many studies have been conducted to predict lane-change intention (Song & Li, 2021).

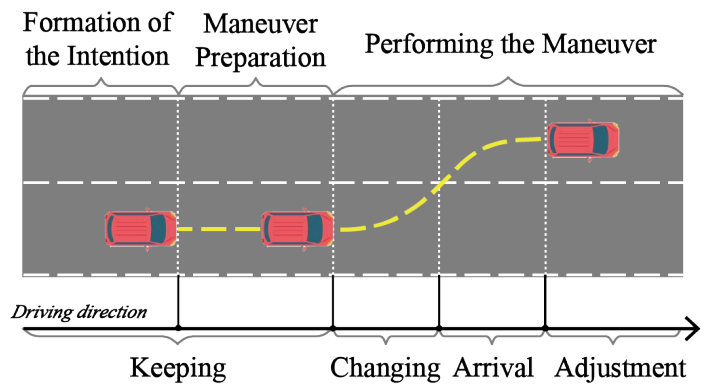

**Fig. 1. Illustration of a typical lane change process.**

The lane-changing behaviors can be divided into discretionary lane-change and mandatory lane-change (Zheng, 2014). Drivers in a discretionary lane-change purse benefits such as faster speeds and more comfortable and safer driving environment, while mandatory lane-changing drivers change their lanes to drive along a preset path. As shown in Fig 1, there are two consecutive statuses before a lane-changing behavior, which are the formation of intention and maneuver preparation. During the keeping phase, drivers compare utilities of different lanes and choose the lane with the highest utility. and then drivers will check the surrounding environment to ensure safety during the lane-changing performing phase.

In the term of intention, it means the thoughts that one has before the actions in cognitive psychology (Carruthers, 2007). Accordingly, lane-changing intention towards the execution of a series of future lane-change control actions. The lane-changing intention can be detected during the phases of maneuver preparation. If the model can identify the lane-changing intention before the subject vehicle performing the maneuver, it is usually considered as lane-changing intention 'prediction'. Otherwise, it is usually considered as lane-changing detection. So, lane-changing intention prediction refers to the prediction of the lane-changing behavior based on a set of data of actions or signals before performing the lane-change and the detection of lane-changing behavior is not considered in this paper for concision and accuracy.

Lane-changing intention prediction is critical for many features of autonomous driving or advanced driver-assistance system (ADAS). Because ADAS need to avoid conflicts with not only other objects but also the drivers. As ADAS share the control authorities with the driver, Lane-changing intention prediction enables the ADAS to detect and predict the driver's lane-changing intention and not against the driver's operation. So accurate drivers' intention prediction especially in lane-change will contribute to a smoother and safer transition between the driver and the autonomous vehicle controller.

Recently, with the advancement of traffic detection and video surveillance technologies, a large number of different variables or parameters used as inputs of prediction models have become available for researchers and vehicle design companies. These new data collection methods have an advantage of allowing a direct observation of driver's behavior and provide larger-scale datasets and available features. Consequently, selecting the most critical features as the inputs becomes an increasingly important topic when developing the lane-changing intention prediction system. The objective of feature selection is three-fold: improving the prediction performance of the predictors, providing faster and more cost-effective predictors, and providing a better understanding of the underlying process that generated the data.

There are two approaches commonly used to explore lane-changing intention prediction in different lane-changing situations, i.e., off-road experiments and field (on-road) studies. Off-road experiments (such as driving simulators and motion tracking devices) have been a mature approach to test how to detect or predict lane-changing intention by capturing the features from the driving behaviors and psychology (Guo et al., 2021; Xing et al., 2019; Deng et al., 2018; Wissing et al., 2017).

Field studies about lane-changes are limited because real vehicle tests on road is costly and may result in safety issues. The nature of field studies is to collect real data of driving behaviors. It requires the research group to have their own test vehicles equipped with all the necessary sensors, data acquisition, and processing systems (Song & Li, 2021). An alternative option is to use publicly available real-world datasets, such as Next Generation Simulation (NGSIM), HighD, PREdiction of vehicles intentions (PREVENTION), Safety Pilot Model Deployment Data (SPMD) (Wang et al., 2019), and Strategic Highway Research Program (SHRP2) (Mahajan et al., 2020; TANG et al., 2020; Deng et al., 2019; Mänttäri et al., 2018; Zhang et al., 2018; Bakhit et al., 2017; Dou et al., 2016; Gao et al., 2021; Wang et al., 2016; Griesbach et al., 2021; Gao et al., 2021; Leonhardt, V., & Wanielik, G., 2017; Bakhit et al., 2017; Li, K., 2016; Gao et al., 2021; Yan et al., 2019; Peng et al., 2015; Deng et al., 2019). They include naturalistic driving data (NDS), GPS data and trajectory data, which own the features of unobtrusive data collection, large-scale of the sample, and real-time continuous dataset comparing with traditional field studies methods (Fan et al., 2019). Both NGSIM and HighD are microscopic vehicle trajectory data, which is widely used in many research, extracted from traffic videos. The trajectory data include time, frame ID, vehicle's longitudinal position and lateral position, velocity, acceleration, preceding and following vehicle ID, time headway, lane number, and other information. NGSIM was collected on Interstate Highway 80 (I-80) and US Route 101 (US-101) in Emeryville, California (FHWA, 2006). And HighD includes six different locations near Cologne, Germany, which are typical German highways with two or three lanes in each direction (. The collection time of the HighD dataset is from 2017 to 2018 and HighD has less noise in the data compared with NGSIM. PREVENTION dataset, which provides a large number of accurate and detailed annotations of vehicles trajectories, has developed a novel benchmark for the prediction of vehicles intentions. The dataset is collected from 6 sensors of different nature (Lidar, Radar, and cameras), which provides both redundancy and complementarity, using an instrumented vehicle driven with providing the ego vehicle's perspective under naturalistic conditions (Izquierdo et al., 2020).

The primary objective of this study is to explore the effectiveness of different features categories combination in lane-changing intention prediction. In order to answer these questions, studies related to lane-changing intention prediction have been selected. The meta-analysis is used to not only evaluate the effectiveness of different features categories combination in lane-changing intention but also capture heterogeneity, effect size combination, and publication bias.

The remainder of this study is organized as follows. Section 2 describes the relevant studies searching on lane-changing intention prediction and meta-analysis method, followed by the results in Section 3. Section 4 concludes the major findings and provides future research directions.

## 2. DATA AND METHODOLOGY
### 2.1 Literature retrieval method
Meta-analysis is a statistical analysis method for the secondary processing of existing literature. And researchers can get high-quality comprehensive scientific conclusions via meta-analysis without experimentations. In meta-analysis, fully searching in related articles is as much essential as the data in experiments. Thus, A set of strict and clear article selection criteria is applied to ensure the validity and representation of the opinions conclude from the result of meta-analysis.

This study used "Google Scholar" and "Web of Science" to retrieve the literature, and the keyword styles were as following, "lane-change" AND "intention" OR

"behavior" AND "prediction". Additionally, to make sure the quality and representativeness of the cited papers in all of the searched literature, selected papers except published in 2021 and 2020 have more than 5 citations. Meanwhile, the journals that the selected papers published are Q1 or Q2.

During the phases of the formation of intention and maneuver preparation, many various specific input factors can be used to predict the driver's lane-changing intention. To avoid generating too many factors match, the inputs are divided into three groups according to their resources: environment context, driver behavior and psychological signals, the status of the vehicle. What needs explanation is that environment context refers to the moveable traffic (including subject vehicle and surrounding vehicles), traffic signs, weather conditions, etc. And the status of the vehicle Specifically referred to the signals that are available through the Control Area Network (CAN), such as steering wheel angle, indicator and brake/gas pedal position.

## 2.2 Meta-analysis

In this study, the meta-analysis is applied to evaluate the effectiveness of different features categories combination in lane-changing intention prediction based on the selected studies in the selected literature. However, due to the insufficient number of studies related to different input type combinations (such as drivers input type only, the combination of driver factors and environmental factors), the meta-analysis is performed on environmental input only, the status of the vehicle input only, and drivers related input type (including Drivers only, Drivers& environment and Drivers& vehicle).

In the meta-analysis, the statistical weight assigned to each study is obtained by the inverse variance approach:

$$W_i = \frac{1}{se_i^2} \qquad (1)$$

Where, $W_i$ denotes the statistical weight; and $se$ is the stand error of study $i$ (Elvik, 2011).

The weighted mean effectiveness based on a set of estimates is:

$$\overline{ES} = \frac{\sum_{i=1}^{g} ES_i W_i}{\sum_{i=1}^{g} W_i} \qquad (2)$$

Where, $\overline{ES}$ denotes the weighted mean effectiveness; and $g$ is the number of estimates of effect that have been combined. There are two methods related to inverse variance approach, i.e., fixed effects model and random effects model. The former one considers the within-study variance only, and the formula is shown in (1), while the random effects model takes the systematic between-study variance into consideration, as shown in the following equations:

$$W_i^* = \frac{1}{se_i^2 + \sigma_\theta^2} \qquad (3)$$

Where, $W_i^*$ is the statistical weight for the random effects model; and $\sigma_\theta^2$ is the between-study variance and calculation process as follows:

$$\sigma_\theta^2 = \frac{Q - g + 1}{\sum W - \left(\sum \frac{W^2}{\sum W}\right)} \tag{4}$$

Where, the $Q$ is the homogeneity test statistic is defined as:

$$Q = \sum_{i=1}^{g} W_i Y_i^2 - \frac{\sum_{i=1}^{g} W_i Y_i}{\sum_{i=1}^{g} W_i} \tag{5}$$

Where, $Y_i$ is the observed effect of the study $i$ (Lipsey and Wilson, 2001). The heterogeneity was tested by $I^2$ test, which is commonly accepted by its higher reliability than the $Q$ test, and $I^2$ could be derived by $Q$ (Higgins and Thompson, 2002; Higgins et al., 2003):

$$I^2 = \begin{cases} \frac{Q - g + 1}{Q} \times 100\%, Q > g - 1 \\ 0, Q < g - 1 \end{cases} \tag{6}$$

Where, $I^2$ is heterogeneity index, $I^2 > 0.75$, there is a large heterogeneity between studies; $0.50 < I^2 < 0.75$ represents a moderate heterogeneity; and if $I^2 < 0.50$, it shows a low heterogeneity within an acceptable range. A random effect model is usually employed when significant heterogeneity exists (Elvik, 2011). In this study, since the influential factors are variable and a random-effects model will cause wider confidence intervals for the estimate results, the random effect model is adopted (Elvik, 2005). The confidence interval at 95% was calculated from (7):

$$95\%\text{CI} = \text{ES} \pm 1.96 \times \frac{1}{\sum_{i=1}^{g} W_i} \tag{7}$$

Furthermore, this research used funnel plot to obtain the presence and degree of publication bias. And the trim-and-fill method was conducted to reduce the publication bias which can improve the accuracy and effectiveness of the estimated result.

**2.2.1. Prediciton performance rate (PPR)**

In this study, the results extracted from the selected papers have different quantitative standards in the evaluation of prediction performance, which makes data various and heterogeneous. Therefore, a new assessment indicator, prediction performance rate (PPR), was developed. By formula (8), the prediction performance could be normalized and universalized for different studies.

$$\text{PPR} = \frac{\sum_{j=1}^{n} ACC_i}{n} - 0.5 \tag{8}$$

Where, $ACC_i$ means the accuracy of prediction of behavior $i$; $n$ is the total number of predicted behavior types. Predicted behavior types usually are divided into lane-keep (LK) and lane-change (LC). Some researchers also consider the difference between turning left (LL) and turning right (LR).

## 3. RESULT AND DISCUSSION

The machine learning and deep learning have been the most popular method in driver's lane-changing intention prediction in the existing literature. And due to the different parameters, structure, even specific factors settled in the model, strong heterogeneity will found in different studies. So it is important to eliminate or reduce heterogeneity is the focus in meta-analysis papers. This study adopted random effects model, which is the most commonly used method to fit the random effect for meta-analysis (Jackson et al., 2010).

Fig. 2 is funnel plots, which show the publication bias among the selected paper. The horizontal axis represents the prediction performance rate under different input types. Values greater than 0 indicate a reduction of prediction performance rate. The vertical axis denotes the stand error of each study. The funnel plot presented an inverted symmetric funnel shape, and the points were distributed symmetrically around the estimated true values of the independent study effect points, representing no bias in the included studies. If bias existed, the funnel plot would be asymmetry and the more significant the asymmetry, the greater the degree of bias. To objectively describe the symmetry of funnel plots of different input types, the rank correlation test was adopted and the results were shown in Table 1.

Table 1. The result of rank correlation test for publication bias.

| INPUT TYPE | Number of analyses | P (Begg' test) | P (Egg' s test) |
|---|---|---|---|
| Environment | 11 | 0.640 | 0.640 |
| Vehicle | 7 | 0.764 | 0.764 |
| Drivers related | 8 | 1.000 | 1.000 |

Table 2. Summary of the cited papers.

| Author (Year) | Sample | | Input Factors | | Algorithm | Accuracy | Predict Horizon |
|---|---|---|---|---|---|---|---|
| | No. Subjects | No. Lane-change | Resource | Type | | | |
| Mahajan et al., 2020 | 334 vehicles | 726 | High D | | SVM | 99%LC and 100% LK | |
| TANG et al., 2020 | | 2231 (1125 LK; 919 LF;187LR) | Trajectory Data | | LSTM | 83.75% | 2s |
| Deng et al., 2019 | 10 drivers | 28 | ENV | | SVM(对于 ENV) | 89.16%LK 95.51%LF 92.69%LR | |
| Mänttäri et al., 2018 | 16400 samples | 3280 | NGSIM I80 | Environment | RNN | 96% | 0.3s |
| Zhang et al., 2018 | 51070 samples | 538 | NGSIM I80 | | GMM-HMM | 81.32% 88.29% 82.49% 85.61% | |
| | 51285 samples | 552 | NGSIM US101 | | | | |
| Bakhit et al., 2017 | 16223 samples | 4600 | Weaving freeway segment in Arlington, Virginia | | ANN | 82.3% | |
| Dou et al., 2016 | | 754 | NGSIM I80 | | Combined Bayesian and Decision Tre | 94.3% | |
| Wissing et al., 2017 | 25 drivers | 343 simulator (181LCL; 162LCR); 89 real world (37LCL; 52LCR) | Simulator; NDS | | QRF | 85%LCL 83%LCR | |
| Gao et al., 2021 | 6 drivers | 674LC(348LF and 326LR, 350 LK) | Simulator | Driver | (LCNet) Joint learning | 82.4% | |
| Wang et al., 2016 | | 440 | Simulator | | MTS-DeepNet | 71.18% | |

Table 2. Summary of the cited papers. (Continue)

| Author (Year) | Sample | | Input Factors | | Algorithm | Accuracy | Predict Horizon |
|---|---|---|---|---|---|---|---|
| | No. Subjects | No. Lane-change | Resource | Type | | | |
| Griesbach et al., 2021 | 57 drivers | 1748(1019LL & 729 LR) | Naturalistic driving data | Vehicle | RNN | 94%LL & 100LR | 2.88s LL & 3.3s LR |
| Gao et al., 2021 | 6 drivers | 674LC (348LL & 326LR, 350 LK) | Video images and physiological | | (LCNet) Joint learning | 85.68% | |
| Deng et al., 2018 | 9 drivers | 116 | camera, radar, lidar, ultrasonics | | HMM | 87.50% | |
| Leonhardt, V., & Wanielik, G., 2017 | 60 drivers | 1869 | NDS | | artificial neural network | 99.10% | 2s |
| Bakhit et al., 2017 | 16223 samples | 4600 | NDS | | ANN | 82.30% | |
| Li, K., 2016 | 50 drivers | 642 | NDS | | HMM-BF | 90.3%LL 93.5%LR | 0.5s |
| Gao et al., 2021 | 6 drivers | 674LC (348LL & 326LR, 350 LK) | Vehicle test | Driver and Vehicle | (LCNet) Joint learning | 88.57 | |
| Guo et al., 2021 | 25 drivers | 1220 ( 620LK & 600 LC) | Simulator | | LSTM | 93.33 | 3 |
| Yan et al., 2019 | 9 drivers | 236LR 196LF | NDS; Physiological signal | | GRU | 94.76% | 0.58s |
| Peng et al., 2015 | 807 samples | 406 | Eye tracking | | BP network | 95.63% | 0 |
| Deng et al., 2019 | 10 drivers | 28 | NDS; Physiological signal | Driver & Environment | RF | 99.90%LK 99.94%LL 99.92%LR | |

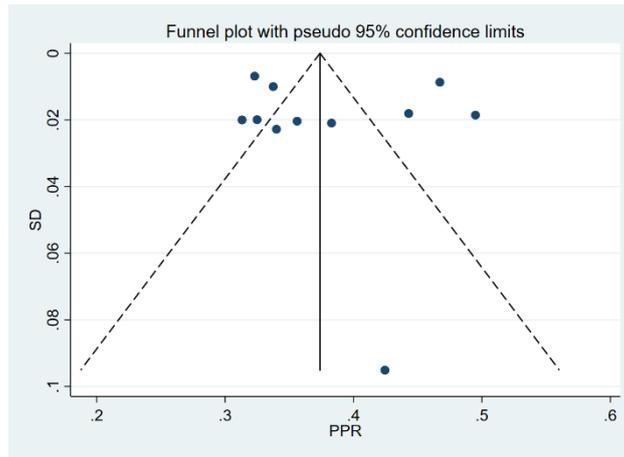

**(a) Environment input type**

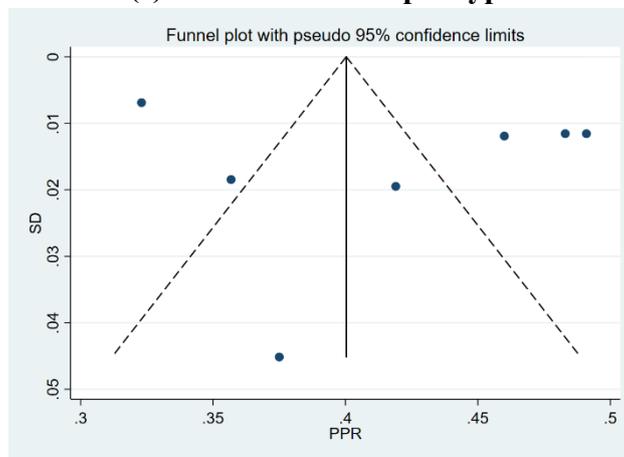

**(b) Vehicle input type**

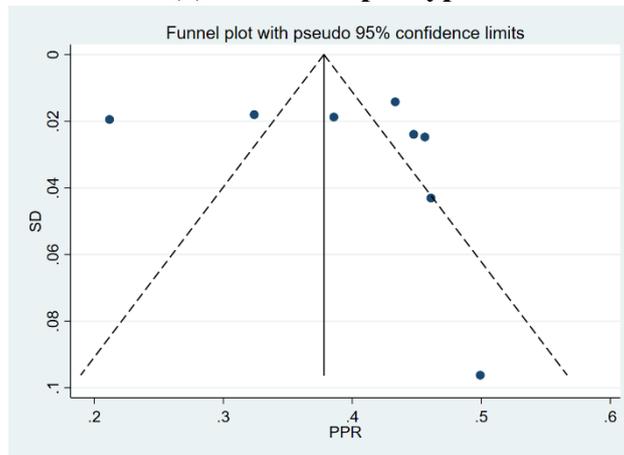

**(c) Driver related input type**

**Fig. 2. Funnel Plots**

The results of rank correlation test indicate that all the funnel plots are symmetric. Because the p-value is larger than 0.05, which means symmetric. Hence, the funnel plots show that the selected research does not have a certain publication bias. The trim-and-fill method was used to mitigate the bias and the meta-analysis results by random effect model as shown in Table 3.

Table 3. Estimated Effect Size of Random Effect Model and Trim-and-fill.

| INPUT TYPE | Random Effect Model | | | After Trim-and-fill | | |
|---|---|---|---|---|---|---|
| | Estimate | 95%CI | $I^2$ | Estimate | 95%CI | $I^2$ |
| Environment | 0.380 | [0.336,0.425] | 96.1% | 0.380 | [0.336,0.425] | 96.1% |
| Vehicle | 0.417 | [0.352,0.482] | 97.8% | 0.417 | [0.352,0.482] | 97.8% |
| Drivers related | 0.394 | [0.328,0.460] | 94.2% | 0.387 | [0.323,0.450] | 94.3% |

As shown in Table 3, Vehicle input type has a better performance in lane-changing intention prediction, compared with environment or even driver combination input type. The possible reason is that the predict horizon has been not considered in prediction performance yet, and the features of vehicle movement has advantage in detecting driving behavior. And strong heterogeneity is found among different studies.

## 4. CONCLUSIONS

This study focused on exploring the effectiveness of different features categories combination in lane-changing intention prediction. Studies related to lane-changing intention prediction have been selected followed by strict standards. The meta-analysis was employed to not only evaluate the effectiveness of different features categories combination in lane-changing intention but also capture heterogeneity, effect size combination, and publication bias. According to the result of the meta-analysis and reviewed research papers, the major conclusions are as follows:

(1) Using input features from different types can lead to different performances.
(2) Vehicle input type has a better performance in lane-changing intention, prediction, compared with environment or even driver combination input type.

In the future, human factors (such as driver character, driving style) will be a significant role in lane-changing prediction. Because different drivers have different driving styles, and the intention inference system cannot work uniformly. Nowadays, most of the prediction work that takes into account the features of the driver is mainly based on physiological signals and behaviors and conducted by simulators. But it is hard to collect plenty of physiological signals and behaviors in real-world due to the cost and privacy policy. Therefore, I think, driving style may play a more important role than other features of the driver. And few researchers pay attention to the data fusion. Considering the limitation of power and computer ability, more variables do not mean better prediction performance. So, the fuzzy data fusion based on the driving style will be meaningful and worth exploring.

# REFERENCES

Ahn, S., Laval, J. (2010). Cassidy, M.J. Effects of merging and diverging on freeway traffic oscillations. Transportation Research Record, (2188), pp. 1-8.

Bakhit, P. R., Osman, O. A., & Ishak, S. (2017). Detecting imminent lane change maneuvers in connected vehicle environments. Transportation Research Record, 2645(1), 168-175.

Bocklisch, F., Bocklisch, S. F., Beggiato, M., & Krems, J. F. (2017). Adaptive fuzzy pattern classification for the online detection of driver lane change intention. Neurocomputing, 262, 148-158.

Carruthers, P. (2007). The illusion of conscious will. Synthese, 159(2), 197-213.

Deng, Q., & Soeffker, D. (2021). A Review of the current HMM-based Approaches of Driving Behaviors Recognition and Prediction. IEEE Transactions on Intelligent Vehicles.

Deng, Q., Wang, J., & Soffker, D. (2018, June). Prediction of human driver behaviors based on an improved HMM approach. In 2018 IEEE Intelligent Vehicles Symposium (IV) (pp. 2066-2071). IEEE.

Elvik, R. (2005). Introductory guide to systematic reviews and meta-analysis. Transportation research record, 1908(1), 230-235.

Elvik, R. (2011). Effects of mobile phone use on accident risk: Problems of meta-analysis when studies are few and bad. Transportation research record, 2236(1), 20-26.

FHWA, 2006. Next Generation Simulation (NGSIM).

Gao, J., Yi, J., & Murphey, Y. L. (2021). Joint Learning of Video Images and Physiological Signals for Lane-Changing Behavior Prediction. Transportmetrica A: Transport Science, (just-accepted), 1-13.

Gao, J., Zhu, H., & Murphey, Y. L. (2019, May). A personalized model for driver lane-changing behavior prediction using deep neural network. In 2019 2nd International Conference on Artificial Intelligence and Big Data (ICAIBD) (pp. 90-96). IEEE.

Griesbach, K., Beggiato, M., & Hoffmann, K. H. (2021). Lane change prediction with an echo state network and recurrent neural network in the urban area. IEEE Transactions on Intelligent Transportation Systems.Song, R., & Li, B. (2021). Surrounding Vehicles' Lane Change Maneuver Prediction and Detection for Intelligent Vehicles: A Comprehensive Review. IEEE Transactions on Intelligent Transportation Systems.

Guo, Y., Zhang, H., Wang, C., Sun, Q., & Li, W. (2021). Driver lane change intention recognition in the connected environment. Physica A: Statistical Mechanics and its Applications, 575, 126057.

Han, T., Jing, J., & Özgüner, Ü. (2019, June). Driving intention recognition and lane change prediction on the highway. In 2019 IEEE Intelligent Vehicles Symposium (IV) (pp. 957-962). IEEE.

Higgins, J. P., & Thompson, S. G. (2002). Quantifying heterogeneity in a meta-analysis. Statistics in medicine, 21(11), 1539-1558.

Higgins, J. P., Thompson, S. G., Deeks, J. J., & Altman, D. G. (2003). Measuring inconsistency in meta-analyses. Bmj, 327(7414), 557-560.

issing, Y., Yan, F., & Feng, D. (2016, July). Lane changing prediction at highway lane drops using support vector machine and artificial neural network classifiers. In 2016 IEEE International Conference on Advanced Intelligent Mechatronics (AIM) (pp. 901-906). IEEE.

Izquierdo, R., Quintanar, A., Parra, I., Fernández-Llorca, D., & Sotelo, M. A. (2019). The prevention dataset: a novel benchmark for prediction of vehicles intentions. In 2019 IEEE Intelligent Transportation Systems Conference (ITSC) (pp. 3114-3121). IEEE.

Izquierdo, R., Quintanar, A., Parra, I., Fernández-Llorca, D., & Sotelo, M. A. (2019, October). The prevention dataset: a novel benchmark for prediction of vehicles intentions. In 2019 IEEE Intelligent Transportation Systems Conference (ITSC) (pp. 3114-3121). IEEE.

Jackson, D., Bowden, J., & Baker, R. (2010). How does the DerSimonian and Laird procedure for random effects meta-analysis compare with its more efficient but harder to compute counterparts?. Journal of Statistical Planning and Inference, 140(4), 961-970.

Kim, I. H., Bong, J. H., Park, J., & Park, S. (2017). Prediction of driver's intention of lane change by augmenting sensor information using machine learning techniques. Sensors, 17(6), 1350.

Leonhardt, V., & Wanielik, G. (2017, July). Feature evaluation for lane change prediction based on driving situation and driver behavior. In 2017 20th International Conference on Information Fusion (Fusion) (pp. 1-7). IEEE.

Leonhardt, V., & Wanielik, G. (2017, October). Neural network for lane change prediction assessing driving situation, driver behavior and vehicle movement. In 2017 IEEE 20th International Conference on Intelligent Transportation Systems (ITSC) (pp. 1-6). IEEE.

Leonhardt, V., & Wanielik, G. (2018). Recognition of lane change intentions fusing features of driving situation, driver behavior, and vehicle movement by means of neural networks. In Advanced Microsystems for Automotive Applications 2017 (pp. 59-69). Springer, Cham.

Li, K., Wang, X., Xu, Y., & Wang, J. (2016). Lane changing intention recognition based on speech recognition models. Transportation research part C: emerging technologies, 69, 497-514.

Lipsey, M. W., & Wilson, D. B. (2001). Practical meta-analysis. SAGE publications, Inc.

Mahajan, V., Katrakazas, C., & Antoniou, C. (2020). Prediction of lane-changing maneuvers with automatic labeling and deep learning. Transportation research record, 2674(7), 336-347.Das, A., Khan, M. N., & Ahmed, M. M. (2020). Detecting lane change maneuvers using SHRP2 naturalistic driving data: a comparative study machine learning techniques. Accident Analysis & Prevention, 142, 105578.

Mänttär, Q., Wang, J., Hillebrand, K., Benjamin, C. R., & Söffker, D. (2019). Prediction performance of lane changing behaviors: a study of combining environmental and eye-tracking data in a driving simulator. IEEE Transactions on Intelligent Transportation Systems, 21(8), 3561-3570.

Mänttäri, J., Folkesson, J., & Ward, E. (2018, June). Learning to predict lane changes in highway scenarios using dynamic filters on a generic traffic representation. In 2018 IEEE Intelligent Vehicles Symposium (IV) (pp. 1385-1392). IEEE.

NHTSA. (2019). Traffic Safety Facts Annual Report Tables 2019. U.S. 26 Department of Transportation.

Pande, A., & Abdel-Aty, M. (2006). Assessment of freeway traffic parameters leading to lane-change related collisions. Accident Analysis & Prevention, 38(5), 936–948.

Peng, J., Guo, Y., Fu, R., Yuan, W., & Wang, C. (2015). Multi-parameter prediction of drivers' lane-changing behaviour with neural network model. Applied ergonomics, 50, 207-217.

Peng, J., Guo, Y., Fu, R., Yuan, W., & Wang, C. (2015). Multi-parameter prediction of drivers' lane-changing behaviour with neural network model. Applied ergonomics, 50, 207-217.

Schlechtriemen, J., Wirthmueller, F., Wedel, A., Breuel, G., & Kuhnert, K. D. (2015, June). When will it change the lane? A probabilistic regression approach for rarely occurring events. In 2015 IEEE Intelligent Vehicles Symposium (IV) (pp. 1373-1379). IEEE.

Singh, S. (2015). Critical reasons for crashes investigated in the national motor vehicle crash causation survey

Song, R. (2021). Driver intention prediction using model-added Bayesian network. Proceedings of the Institution of Mechanical Engineers, Part D: Journal of Automobile Engineering, 235(5), 1236-1244.

Song, R., & Li, B. (2021). Surrounding Vehicles' Lane Change Maneuver Prediction and Detection for Intelligent Vehicles: A Comprehensive Review. IEEE Transactions on Intelligent Transportation Systems.

Tang, J., Liu, F., Zhang, W., Ke, R., & Zou, Y. (2018). Lane-changes prediction based on adaptive fuzzy neural network. Expert Systems with Applications, 91, 452-463.

Tang, L., Wang, H., Zhang, W., Mei, Z., & Li, L. (2020). Driver lane change intention recognition of intelligent vehicle based on long short-term memory network. IEEE Access, 8, 136898-136905.

Wang, C., Delport, J., & Wang, Y. (2019). Lateral motion prediction of on-road preceding vehicles: a data-driven approach. Sensors, 19(9), 2111.

Wang, X., Murphey, Y. L., & Kochhar, D. S. (2016, July). MTS-DeepNet for lane change prediction. In 2016 International Joint Conference on Neural Networks (IJCNN) (pp. 4571-4578). IEEE.

Wissing, C., Nattermann, T., Glander, K. H., & Bertram, T. (2017, June). Probabilistic time-to-lane-change prediction on highways. In 2017 IEEE Intelligent Vehicles Symposium (IV) (pp. 1452-1457). IEEE.

Wissing, C., Nattermann, T., Glander, K. H., Hass, C., & Bertram, T. (2017). Lane change prediction by combining movement and situation based probabilities. IFAC-PapersOnLine, 50(1), 3554-3559.

Woo, H., Ji, Y., Kono, H., Tamura, Y., Kuroda, Y., Sugano, T., ... & Asama, H. (2016). Automatic detection method of lane-changing intentions based on relationship with adjacent vehicles using artificial potential fields. International Journal of Automotive Engineering, 7(4), 127-134.

Woo, H., Ji, Y., Kono, H., Tamura, Y., Kuroda, Y., Sugano, T., ... & Asama, H. (2016, October). Dynamic potential-model-based feature for lane change prediction. In 2016 IEEE International Conference on Systems, Man, and Cybernetics (SMC) (pp. 000838-000843). IEEE.

Xing, Y., Lv, C., Wang, H., Cao, D., & Velenis, E. (2020). An ensemble deep learning approach for driver lane change intention inference. Transportation Research Part C: Emerging Technologies, 115, 102615.

Xing, Y., Lv, C., Wang, H., Wang, H., Ai, Y., Cao, D., ... & Wang, F. Y. (2019). Driver lane change intention inference for intelligent vehicles: framework, survey, and challenges. IEEE Transactions on Vehicular Technology, 68(5), 4377-4390.

Yan, Z., Yang, K., Wang, Z., Yang, B., Kaizuka, T., & Nakano, K. (2019, June). Time to lane change and completion prediction based on Gated Recurrent Unit Network. In 2019 IEEE Intelligent Vehicles Symposium (IV) (pp. 102-107). IEEE.

Zhang, Y., Lin, Q., Wang, J., Verwer, S., & Dolan, J. M. (2018). Lane-change intention estimation for car-following control in autonomous driving. IEEE Transactions on Intelligent Vehicles, 3(3), 276-286.

Zheng, Z. (2014). Recent developments and research needs in modeling lane changing. Transportation Research Part B 60, 16–32.